\documentclass[twoside,11pt]{article}

\usepackage[abbrvbib,preprint]{jmlr2e}

\usepackage{jmlr2e}
\usepackage{graphicx}
\usepackage{floatrow}
\newfloatcommand{capbtabbox}{table}[\captop][\FBwidth]
\usepackage{wrapfig}
\usepackage{natbib}
\usepackage{amsmath}
\usepackage{authblk}

\usepackage{authblk}
\usepackage{wrapfig}
\usepackage{lipsum} 

\usepackage{lastpage}

\jmlrheading{25}{2024}{1-\pageref{LastPage}}{6/24; Revised 9/24}{10/24}{24-0991}{Mirco Ravanelli, Titouan Parcollet, et al.}

\ShortHeadings{Open-Source Conversational AI with SpeechBrain 1.0}{Ravanelli, Parcollet, et al.}

\firstpageno{1}

\makeatletter
\let\@fnsymbol\@alph
\makeatother

\begin{document}

\title{Open-Source Conversational AI with SpeechBrain 1.0}



\author{
  \small
  \textbf{Mirco Ravanelli}\textsuperscript{1,2,5}, 
  \textbf{Titouan Parcollet}\textsuperscript{4,6}, 
  \textbf{Adel Moumen}\textsuperscript{3},
  \textbf{Sylvain de Langen}\textsuperscript{3},
  \textbf{Cem Subakan}\textsuperscript{7,2,1},
  \textbf{Peter Plantinga}\textsuperscript{2},
  \textbf{Yingzhi Wang}\textsuperscript{8},
  \textbf{Pooneh Mousavi}\textsuperscript{1,2}, 
  \textbf{Luca Della Libera}\textsuperscript{1,2},
  \textbf{Artem Ploujnikov}\textsuperscript{5,2},
  \textbf{Francesco Paissan}\textsuperscript{9,14},
  \textbf{Davide Borra}\textsuperscript{10}, 
  \textbf{Salah Zaiem}\textsuperscript{11},
  \textbf{Zeyu Zhao}\textsuperscript{12},
  \textbf{Shucong Zhang}\textsuperscript{4},
  \textbf{Georgios Karakasidis}\textsuperscript{12},
  \textbf{Sung-Lin Yeh}\textsuperscript{12},
  \textbf{Pierre Champion}\textsuperscript{13},
  \textbf{Aku Rouhe}\textsuperscript{14,18},
  \textbf{Rudolf Braun}\textsuperscript{20},
  \textbf{Florian Mai}\textsuperscript{19},
  \textbf{Juan Zuluaga-Gomez}\textsuperscript{20,21},
  \textbf{Seyed Mahed Mousavi}\textsuperscript{15},
  \textbf{Andreas Nautsch}\textsuperscript{3},
  \textbf{Ha Nguyen}\textsuperscript{3},
  \textbf{Xuechen Liu}\textsuperscript{17},
  \textbf{Sangeet Sagar}\textsuperscript{16},
  \textbf{Jarod Duret}\textsuperscript{3},
  \textbf{Salima Mdhaffar}\textsuperscript{3},
  \textbf{Gaëlle Laperrière}\textsuperscript{3},
  \textbf{Mickael Rouvier}\textsuperscript{3},
  \textbf{Renato De Mori}\textsuperscript{3,22},
  \textbf{Yannick Estève}\textsuperscript{3} \\
  \textsuperscript{1}Concordia University,
  \textsuperscript{2}Mila-Quebec AI Institute,
  \textsuperscript{3}Avignon University,
  \textsuperscript{4}Samsung AI Center Cambridge,
  \textsuperscript{5}Université de Montréal,
  \textsuperscript{6}University of Cambridge,
  \textsuperscript{7}Laval University,
  \textsuperscript{8}Zaion,
  \textsuperscript{9}Fondazione Bruno Kessler,
  \textsuperscript{10}University of Bologna,
  \textsuperscript{11}Telecom Paris,
  \textsuperscript{12}University of Edinburgh,
  \textsuperscript{13}Inria,
  \textsuperscript{14}Aalto University,
  \textsuperscript{15}University of Trento,
  \textsuperscript{16}Saarland University,
  \textsuperscript{17}National Institute of Informatics - Tokyo,
  \textsuperscript{18}Silo AI,
  \textsuperscript{19}KU Leuven,
  \textsuperscript{20}Idiap,
  \textsuperscript{21}EPFL,
  \textsuperscript{22}McGill University
}

%


\editor{Alexandre Gramfort}

\maketitle

\begin{abstract}
SpeechBrain\footnote{\url{https://speechbrain.github.io/}} is an open-source Conversational AI toolkit based on PyTorch, focused particularly on speech processing tasks such as speech recognition, speech enhancement, speaker recognition, text-to-speech, and much more.
It promotes transparency and replicability by releasing both the pre-trained models and the complete \textit{``recipes''} of code and algorithms required for training them.
This paper presents SpeechBrain 1.0, a significant milestone in the evolution of the toolkit, which now has over 200 recipes for speech, audio, and language processing tasks, and more than 100 models available on Hugging Face.
SpeechBrain 1.0 introduces new technologies to support diverse learning modalities, Large Language Model (LLM) integration, and advanced decoding strategies, along with novel models, tasks, and modalities. It also includes a new benchmark repository, offering researchers a unified platform for evaluating models across diverse tasks.

\end{abstract}

\begin{keywords}
  Conversational AI, open-source, speech processing, deep learning.
\end{keywords}

\section{Introduction}
Conversational AI is experiencing extraordinary progress, with Large Language Models (LLMs) and speech assistants rapidly evolving and becoming widely adopted in the daily lives of millions of users ~\citep{mctear2021conversational}.
However, this quick evolution poses a challenge to a fundamental pillar of science: \textit{reproducibility}.
Replicating recent findings is often difficult or impossible for many researchers due to limited access to data, computational resources, or code ~\citep{kapoor_leakage_2023}. 
The open-source community is making a remarkable collective effort to mitigate this \textit{``reproducibility crisis"}, yet many contributors primarily release pre-trained models only,  known as open-weight \citep{openweight}. While this is a step forward, it is still very common for the data and algorithms used to train them to remain undisclosed.
We helped address this problem by releasing SpeechBrain~\citep{speechbrain}, a PyTorch-based open-source toolkit designed for accelerating research in speech, audio, and text processing. 
We ensure replicability by releasing pre-trained models for various tasks and providing the \textit{``recipe"} for training them from scratch, conveniently including all necessary algorithms and code. A few other open-source toolkits, like NeMo~\citep{nemo2019} and ESPnet~\citep{watanabe2018espnet}, also support multiple Conversational AI tasks, each excelling in different applications. A more detailed discussion of the related toolkits can be found in Appendix A.

This paper introduces SpeechBrain 1.0, a remarkable milestone resulting from years of collaboration between the core development team and our community volunteers. We will outline key technical updates for supporting novel learning methods, LLM integration, advanced decoding strategies, new models, tasks, and modalities. We also present a new benchmark repository designed to facilitate model comparisons across tasks.


\begin{wrapfigure}{l}{8.5cm}
    \centering
    
\includegraphics[width=8.5cm]{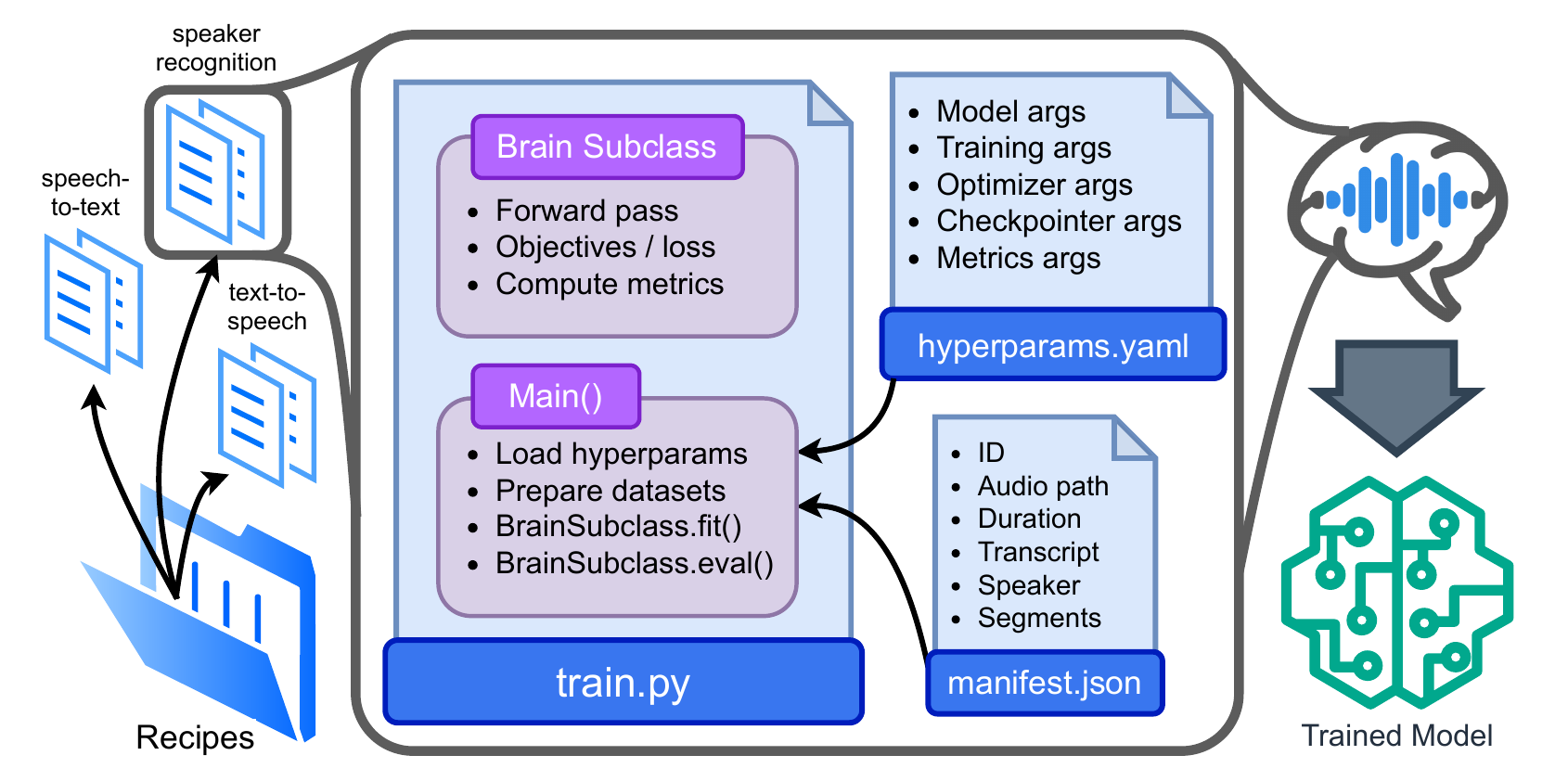}
\caption{SpeechBrain architecture overview.}
\end{wrapfigure}


\section{Overview of SpeechBrain}
Since its launch in March 2021, SpeechBrain has grown rapidly and emerged as one of the most popular toolkits for speech processing. It is downloaded 2.5 million times monthly, used in 2200 repositories, has 8.6k GitHub stars, and 154 contributors.
Despite its constant evolution, we remain faithful to the original design principles. 
We prioritized \textit{replicability} by releasing both training recipes and pre-trained models. Moreover, 95\% of our recipes utilize freely available data and include comprehensive training logs, checkpoints, and other essential information. 
We made SpeechBrain \textit{easy to use} by providing comprehensive documentation, examples, and tutorials. Our modular architecture facilitates easy integration or modification of modules. We built it on PyTorch standard interfaces (e.g., \texttt{torch.nn.Module}, \texttt{torch.optim}, \texttt{torch.utils.data.Dataset}), enabling seamless integration with the PyTorch ecosystem~\citep{sb2022tutorial}.
It is released under the Apache 2.0 license.

\subsection{Architecture Overview}
Training a model with SpeechBrain involves combining the \textit{training script}, the \textit{hyperparameter} file, and the \textit{data manifest} files, as depicted in Figure 1. 
First, users need to specify the data for training, validation, and testing using CSV or JSON files. These formats are supported because they allow flexible and intuitive declaration of input files and annotations.
Next, users must design a model and define its hyperparameters using a modified YAML format known as HyperPyYAML. This format facilitates complex yet elegant parameter configurations, defining objects and their associated arguments.
Finally, users write the training script, which orchestrates all the steps to train the model. The training procedure is integrated into a single Python script which utilizes a specialized \texttt{Brain} class designed to make the process intuitive and standardized.
Our toolkit natively implements popular models, efficient sequence-to-sequence learning, data handling, distributed training, beam search decoding, evaluation metrics, and data augmentation, across over 200 training recipes for widely used research datasets and more than 100 pretrained models.

\begin{table*}[t!]
\scriptsize
\centering
\caption{Summary of the technology supported by SpeechBrain 1.0.}
\label{tab:techniques}
\resizebox{\textwidth}{!}{%
\begin{tabular}{|l|p{0.85\linewidth}|}
\hline
\textbf{Modality} & \textbf{Task and Techniques} \\ \hline
Audio & Vocoding, Audio Augmentation, Feature Extraction, Sound Event Detection, Beamforming. \\ \hline
Speech & Speech Recognition, Enhancement, Separation, Text-to-Speech, Speaker Recognition, Speech-to-Speech Translation, Spoken Language Understanding, Voice Activity Detection, Diarization, Emotion Recognition, Emotion Diarization, Language Identification, Self-Supervised Training, Metric Learning, Forced Alignment. \\ \hline
Text & LM Training, LLM Fine-Tuning, Dialogue Modeling, Response Generation, Grapheme-to-Phoneme. \\ \hline
EEG & Motor Imagery, P300, SSVEP Classification. \\ \hline
\end{tabular}%
}
\end{table*}


\section{Recent Developments}
SpeechBrain now supports a wide array of tasks. Please, refer to Table 1 for a complete list as of October 2024. The main improvements in SpeechBrain 1.0 include:

\begin{itemize}

\item \textbf{Learning Modalities}: We expanded the support for emerging deep learning modalities. For continual learning, we implemented methods like \textit{Rehearsal}, \textit{Architecture}, and \textit{Regularization}-based approaches~\citep{dellalibera2023clmasr}. For interpretability, we developed both post-hoc and design-based methods, including Post-hoc Interpretation via Quantization \citep{paissan2023posthoc}, Listen to Interpret \citep{l2i}, Activation Map Thresholding (AMT) for Focal Networks \citep{dellalibera2024focal}, and Listenable Maps for Audio Classifiers \citep{lmac}. We also implemented audio generation using standard and latent diffusion techniques, along with DiffWave \citep{kong2020diffwave} as a novel vocoder based on diffusion. Lastly, efficient fine-tuning strategies have been introduced for faster inference using speech self-supervised models~\citep{zaiem2023icassp}.
We implemented wav2vec2 SSL pretraining from scratch as described by ~\citep{baevski2020wav2vec}. This enabled efficient training of a 1-billion-parameter SSL model for French on 14,000 hours of speech using over 100 A100 GPUs, showcasing the scalability of SpeechBrain \citep{PARCOLLET2024101622}. We also released the first open-source implementation of the BEST-RQ model ~\citep{bestrq}.
\item \textbf{Models and Tasks}: We developed several new models and expanded support for various tasks. For speech recognition, we introduced new alternatives to the Transformer architecture like HyperConformer~\citep{mai2023hyperconformer} and Branchformer~\citep{PengDL022}, along with a Streamable Conformer Transducer. We implemented the Stabilised Light Gated Recurrent Units~\citep{sligru}, an improved version of the light GRU for more efficient learning~\citep{ligru}. We now support models for discrete audio tokens (e.g., discrete wav2vec, HuBERT, WavLM, EnCodec, DAC, and Speech Tokenizer), which form the basis for modern multimodal LLMs~\citep{mousavi2024}.
Additionally, we introduced technology for Speech Emotion Diarization~\citep{wang2023speech}. To improve usability and flexibility, we refactored speech augmentation techniques \citep{contamination2, contamination}.
In terms of new modalities, SpeechBrain 1.0 now supports electroencephalographic (EEG) signal processing \citep{SpeechBrainMOABB}. Supporting EEG aligns with our long-term goal of enabling natural human-machine conversation, including for those who cannot speak.
Thanks to deep learning, the technology used for speech and EEG processing is getting similar, simplifying their integration in a single toolkit.
SpeechBrain 1.0 is a step in this direction by supporting EEG tasks such as motor imagery, P300, and SSVEP classification with EEGNet~\citep{lawhern2018}, ShallowConvNet~\citep{schirrmeister2017}, and EEGConformer~\citep{song2023}.


\item \textbf{Decoding Strategies}: We improved beam search algorithms for speech recognition and translation. Our update simplifies code with separate scoring and search functions. This update allows easy integration of various scorers, including n-gram language models and custom heuristics. Additionally, we support pure CTC training, RNN-T latency controlled beamsearch \citep{RNNT-beam}, batch and GPU decoding \citep{kim2017joint}, and N-best hypothesis output with neural language model rescoring \citep{salazar2019masked}. We also offer an interface to Kaldi2 (k2) for search based on Finite State Transducers (FST) \citep{10094567} and KenLM for fast language model rescoring \citep{heafield-2011-kenlm}.


 \item  \textbf{Integration with LLMs}: LLMs are crucial in modern Conversational AI. We enhanced our interfaces with popular models like GPT-2~\citep{radford2019language} and Llama 2/3~\citep{touvron2023llama}, enabling easy fine-tuning for tasks such as dialogue modeling and response generation~\citep{mousavi2024are}. We also implemented LTU-AS \citep{gong2023joint}, a speech LLM designed to jointly understand audio and speech. Additionally, LLMs can be used to rescore n-best hypotheses provided by speech recognizers~\citep{tur2024progres}.

\item \textbf{Benchmarks}: We launched a new benchmark repository for facilitating community standardization across various areas of broad interest. Currently, we host four benchmarks: \textit{CL-MASR} for multilingual ASR continual learning~\citep{dellalibera2023clmasr}, \textit{MP3S} for speech self-supervised models with customizable probing heads~\citep{zaiem2023speech}, \textit{DASB} for discrete audio token assessment \citep{mousavi2024dasb}, and \textit{SpeechBrain-MOABB} \citep{SpeechBrainMOABB}, which is based on MOABB \citep{Aristimunha_Mother_of_all} and MNE \citep{GRAMFORT2014446}, for evaluating EEG models. 
\end{itemize}



\section{Conclusion and Future Work}
We presented SpeechBrain 1.0, a significant advancement in the evolution of the SpeechBrain project. We outlined the main updates, including novel learning modalities, models, tasks, and decoding strategies, alongside our efforts in benchmarking initiatives. For an overview of further improvements, please visit the project website.
Looking ahead, we plan to keep serving our community with advancements on both large-scale, small-footprint, and multi-modal models.
We plan to fully support training multimodal large language models (MLLMs) that integrate text, speech, and audio processing tasks into a single unified foundation model.

\newpage

\section*{Acknowledgment}
We would like to thank our sponsors: HuggingFace, Samsung AI Center Cambridge, Baidu, OVHCloud, ViaDialog, and Naver Labs Europe. A special thank you to all the contributors who made SpeechBrain 1.0 possible. 
We thank the Torchaudio team \citep{hwang2023torchaudio} for helpful discussion and support.
We acknowledge the support of the Natural Sciences and Engineering Research Council of Canada (NSERC), the Digital Research Alliance of Canada (alliancecan.ca), and the Amazon Research Award (ARA).
We also thank Jean Zay GENCI-IDRIS for their support in computing (Grant 2024-A0161015099 and Grant 2022-A0111012991), and the LIAvignon Partnership Chair in AI.






\small
\bibliography{sample}

\newpage
\appendix

\section{Related Toolkits}
Some open-source toolkits for Conversational AI have been developed in recent years, with NeMo\footnote{\url{https://github.com/NVIDIA/NeMo}} \citep{nemo2019} and ESPnet\footnote{\url{https://github.com/espnet/espnet}} being the most relevant for SpeechBrain. While all of these toolkits share the common goal of making Conversational AI more accessible, each is designed with different structures and for specific use cases, meaning the best toolkit to use depends on the particular task and user needs.
NeMo, for instance, is industry-focused, offering ready-to-use solutions, but may provide less flexibility for extensive customization compared to SpeechBrain, which is more research-oriented. ESPnet also supports various tasks with competitive performance, but SpeechBrain stands out for its comprehensive documentation, beginner-friendly tutorials, simplicity, and lightweight design with fewer dependencies.
Another related toolkit is k2\footnote{\url{https://github.com/k2-fsa/k2}}~\citep{10094567}, which integrates Finite State Automaton (FSA) and Finite State Transducer (FST) algorithms into autograd-based machine learning frameworks like PyTorch and TensorFlow. We found these features extremely valuable, so we developed an interface that facilitates the seamless integration of k2 within SpeechBrain.

Beyond general-purpose toolkits for Conversational AI and speech processing, we saw the evolution of more task-specific toolkits. A notable example is pyannote\footnote{\url{https://github.com/pyannote/pyannote-audio}} ~\citep{Bredin23}, which is primarily designed for speaker diarization. It aims to provide effective APIs for specific tasks to serve a broad user base. In contrast, SpeechBrain focuses on advancing research by also offering training recipes.
Lastly, we also have seen the rise of popular speech benchmarks such as SUPERB\footnote{\url{https://superbbenchmark.github.io/}}~\citep{yang21c_interspeech}, which provides a set of resources to evaluate the performance of universal shared representations for speech processing. While SUPERB is highly valuable to the community, SpeechBrain has a broader goal. In addition to benchmarking existing models, we indeed aim to provide all the necessary code to train models from scratch.

For the EEG modality, we rely on two key dependencies: MOABB\footnote{\url{https://github.com/NeuroTechX/moabb}}\citep{Aristimunha_Mother_of_all} and MNE\footnote{\url{https://mne.tools/}}\citep{GRAMFORT2014446}. MOABB is chosen for its user-friendly interface and extensive support for a wide range of EEG datasets, while MNE is used for its comprehensive and standardized data preprocessing pipeline. We also offer an integration with Braindecode\footnote{\url{https://braindecode.org/}}~\citep{HBM:HBM23730}, with a tutorial that explains how to connect it with SpeechBrain.

\begin{table}[t!]
\centering
\caption{Comparison of Equal Error Rated (EER\%) between the original ECAPA-TDNN paper and the SpeechBrain re-implementation.}
\begin{tabular}{|l|c|}
\hline
\textbf{ECAPA-TDNN} & \textbf{EER} \\
\hline
Original Paper & 0.87\% \\
\hline
SpeechBrain & 0.81\% \\
\hline
\end{tabular}
\label{tab:EER_comparison}
\end{table}

\section{Model Replication}
One of the important contributions of SpeechBrain is replicating existing models, which may be closed-source, open-weight only, or models published without accompanying code. This process is often time-consuming and challenging, as successful replication is far from trivial. 

Throughout the project, this replication process has been systematically applied to models not originally developed within SpeechBrain across various tasks, including speaker recognition with ECAPA-TDNN~\citep{ecapa}, speech recognition with Conformers~\citep{conformer} and Branchformers~\citep{branchformer}, speech separation with SkiM~\citep{skim}, DualPath RNN~\citep{dprnn}, and ConvTasNET~\citep{convtasnet}, speech synthesis with Tacotron2~\citep{tacotron}, FastSpeech2~\citep{fastspeech} and HiFi-GAN~\citep{hifigan}, self-supervised learning with Wav2vec2~\citep{wav2vec2}, and BEST-RQ~\citep{bestrq}, and many others.
In all the aforementioned cases, we successfully replicated the models and, in some cases, even improved their performance.

One notable example is the replication of the ECAPA-TDNN model for speaker verification. Through collaboration with the original developers, we released the first open-source version of the model. We not only replicated the results from the original paper but also achieved slight improvements, as detailed in Table \ref{tab:EER_comparison}. The improvement primarily originated from a more robust data augmentation strategy and a more careful selection of the training hyperparameters.

\end{document}